# Non-local Low-rank Cube-based Tensor Factorization for Spectral CT Reconstruction

Weiwen Wu, Fenglin Liu, Yanbo Zhang, *Senior Member, IEEE,* Qian Wang and Hengyong Yu, *Senior Member, IEEE*

*Abstract*—Spectral computed tomography (CT) reconstructs material-dependent attenuation images from the projections of multiple narrow energy windows which is meaningful for material identification and decomposition. Unfortunately, the multi-energy projection datasets usually have lower signal-noise-ratios (SNR). Very recently, a spatial-spectral cube matching frame (SSCMF) was proposed to explore the non-local spatial-spectral similarities for spectral CT. This method constructs a group by clustering up a series of non-local spatial-spectral cubes. The small size of spatial patches for such a group makes the SSCMF fail to fully encode the sparsity and low-rank properties. The hard-thresholding and collaboration filtering in the SSCMF also cause difficulty in recovering the image features and spatial edges. While all the steps are operated on 4-D group, the huge computational cost and memory load might not be affordable in practice. To avoid the above limitations and further improve image quality, we first formulate a non-local cube-based tensor instead of group to encode the sparsity and low-rank properties. Then, as a new regularizer, the Kronecker- Basis-Representation (KBR) tensor factorization is employed into a basic spectral CT reconstruction model to enhance the capability of image feature extraction and spatial edge preservation, generating a non-local low-rank cube-based tensor factorization (NLCTF) method. Finally, the split-Bregman method is adopted to solve the NLCTF model. Both numerical simulations and preclinical mouse studies are performed to validate and evaluate the NLCTF algorithm. The results show that the NLCTF method outperforms other state-of-the-art competing algorithms.

## I. Introduction

THE spectral computed tomography (CT) has obtained a great achievement in terms of tissue characterization [1], lesion detection and material decomposition [2], *etc*. As a special case, the dual-energy CT (DECT) uses two different energy settings to discriminate material components in terms of their energy-related attenuation characteristics [3]. However, the DECT usually employs conventional detectors (i.e., energy-integrating detectors), and its results are often corrupted by beam hardening and spectral blurring. Besides, there are only two different energy source/detector pairs. As a result, only two or three basis material maps can be accurately decomposed. The photon-counting detectors (PCDs) illuminate the prospects of multi-energy CT in practical applications because PCDs can distinguish each incident photon energy by recording pulse height [4]. Theoretically speaking, compared with the conventional energy integrating detector, a PCD can improve signal-to-noise ratio with reduced dose by accounting the number of received photons. However, the PCD has different responses to individual photon's energy. This can lead to spectral distortions, including charge sharing, K-escape, fluorescence x-ray emission and pulse pileups. These distortions can further corrupt the spectral CT projection datasets with complicated noises [5]. Therefore, it is difficult to obtain higher signal-noise-ratio (SNR) projections and reconstruct satisfactory spectral CT images. Alternatively, high quality spectral images can be achieved with higher-powered PCD or superior reconstruction methods [6]. In this work, we mainly focus on improving image quality by developing a more powerful reconstruction algorithm.

Many attempts have been made to reconstruct high quality spectral CT images. According to the employed prior knowledge, in our opinion, all of these efforts can be divided into two categories: empirical-knowledge and prior-image-knowledge based methods [7]. The empirical-knowledge based methods first convert the spectral images into a unified and image-independent transformation domain, and then formulate a sparsity/low-rank reconstruction model of the transform coefficients in terms of an $L_0$-norm, nuclear-norm or $L_1$-norm. Considering the diversity of targets, different empirical-knowledge methods were employed, such as total variation (TV) [8], tensor-based nuclear norm [9], PRISM (prior rank, intensity and sparsity model) [10, 11], tensor PRISM [12, 13], superiorization-based PRISM [14], piecewise linear tight frame transform [15], total nuclear variation (TVN) [16], patch-based low-rank [17], tensor nuclear norm (TNN) with TV [18], structure tensor TV [19], nonlocal low-rank and sparse matrix decomposition [20], multi-energy non-local means (MENLM) [21], spatial spectral nonlocal means [22], *etc*. However, image similarities in non-local spatial space are usually ignored among these methods. Very recently, considering the non-local similarity within spatial-spectral space, we proposed a spatial-spectral cube matching frame (SSCMF) algorithm by stacking up a series of similar small cubes (4 × 4 × 4) to form a 4-D group and then operating hard-thresholding and collaboration filtering on the group [7]. The length of the patches in a group is usually too small to accurately characterize the sparsity and low-rank property. The hard-thresholding and collaboration filtering are rough in image feature recovery and spatial edge

This work was supported in part by the National Natural Science Foundation of China (No. 61471070), National Instrumentation Program of China (No. 2013YQ030629), NIH/NIBIB U01 grant (EB017140) and China Scholarship Council (No. 201706050070).

Wu and Liu* (liufl@cqu.edu.cn) are with the Key Lab of Optoelectronic Technology and Systems, Ministry of Education, Chongqing University, Chongqing 400044, China. Asterisk indicates the corresponding author..

Zhang, Wang and Yu*(E-mail: hengyong-yu@ieee.org) are with the Department of Electrical and Computer Engineering, University of Massachusetts Lowell, Lowell, MA 01854, USA. Asterisk indicates the corresponding author.

preservation. Besides, both the hard-thresholding filtering and collaboration filtering are applied on the formulated 4D group, and it might not be affordable for such huge computational cost and memory load in practice.

The prior-image-knowledge based methods explore both image sparsity and similarity by adopting high quality prior images, such as constructing a redundant dictionary [23]. A dual-dictionary learning (DDL) method was applied to sparse-view spectral CT reconstruction [24]. A tensor dictionary learning (TDL) was introduced to explore the image similarity among different energy bins [25]. Considering the similarity between the image gradient of different energy bins, the image gradient $L_0$-norm was incorporated into the TDL ($L_0$TDL) framework for sparse-view spectral CT reconstruction [26]. The spectral prior image constrained compressed sensing algorithm (spectral PICCS)[27], TV-TV and total variation spectral mean (TV-SM) methods [28] can also be considered as prior-image-knowledge based methods, where a high quality image is treated as prior to constrain the final solution [29]. Very recently, an average-image-incorporated BM3D technology was developed to enhance the correlations among energy bin images [30]. However, the high quality prior images may not be available in practice. In addition, they do not fully utilize the similarities within a single channel.

To handle the aforementioned issues, in this paper, a non-local low-rank cube-based tensor (NLCT) will be constructed to fully explore the similarities and features within the spatial-spectral domain. Compared with the group formulation in the SSCMF algorithm, the NLCT unfolds a 2D spatial image patch as a column vector and the 4-D group degrades to a 3D cube. Tucker [31] and canonical polyadic (CP) [32] are two classic tensor decomposition techniques. Specifically, the Tucker decomposition treats a tensor as an affiliation of the orthogonal bases along all its modes integrated by a core coefficient tensor, and the CP factorizes a tensor as a summation of rank-1 Kronecker bases. However, the CP decomposition cannot characterize well low-rank property of the tensor subspaces along its modes, and the Tucker decomposition usually fails to evaluate tensor sparsity with the volume of core tensor [33, 34]. To address those issues, a Kronecker-Basis-Representation (KBR) measure will be adopted. In 2017, it was first proposed for multispectral image denoising and completion with excellent results [33-35]. Recently, the KBR tensor decomposition was also applied to low-dose dynamic cerebral perfusion CT reconstruction [36].

In this paper, we propose a NLCT factorization (NLCTF) model in terms of KBR regularization for low-dose spectral CT reconstruction. Compared with the previous SSCMF method, the NLCTF formulates cube-based tensor so that it can better explore the non-local spatial and spectral similarity of spectral CT images. The KBR tensor decomposition regularization outperforms the hard-thresholding and collaboration filtering in the SSCMF algorithm in feature extraction, image edge preservation and noise suppression. Our contributions are threefold. First, by considering the characters of spectral CT images, we creatively formulate the non-local low-rank cube-based tensor. Second, we analyze the features of spectral images and employ the KBR regularization term to further exploit the image low-rank and sparsity. Based on the small cube-based 3D low-rank tensors, we establish the NLCTF spectral CT reconstruction model. Third, because of the advantages of the split-Bregman frame [37, 38] for our application in spectral CT, it was employed to solve the NLCTF model rather than the ADMM strategy.

The rest of this paper is organized as follows. In section II, the mathematic model is constructed and the reconstruction method is developed. In section III, numerical simulations and preclinical experiments are designed and performed to validate and evaluate the proposed algorithm. In section IV, some related issues are discussed and conclusions are made.

## II. METHOD

### 2.A. KBR-based Tensor Factorization

A $N^{th}$ order tensor can be denoted as $\mathcal{X} \in \mathcal{R}^{I_1 \times I_2 \times I_3 \times \ldots \times I_N}$. The KBR measure for a tensor $\mathcal{X}$ can be expressed as:

$$m(\mathcal{X}) = \|\mathcal{C}\|_0 + \alpha \prod_{n=1}^{N} rank(X_{(n)}), \quad (1)$$

where $\|.\|_0$ represents the $L_0$ norm, $\mathcal{C} \in \mathcal{R}^{I_1 \times I_2 \times I_3 \times \ldots \times I_N}$ is the core tensor of $\mathcal{X}$ with higher order singular value decomposition (HOSVD), $X_{(n)}$ represents the unfolding matrix with the mode-$n$, and $\alpha > 0$ is a tradeoff parameter to balance the roles of two terms. The first term in (1) constrains the number of Kronecker bases for representing the target tensor, complying with intrinsic mechanism of the CP decomposition [32]. The second term inclines to regularize the low-rank property of the subspace spanned upon each tensor mode, which can be considered as a nonzero-cube in the core tensor space. The KBR measurement facilitates both the inner sparsity of core tensor $\mathcal{C}$ and low-rank property of all tensor unfolding modes $X_{(n)}, n = 1, \ldots, N$. Compared with the conventional tensor sparsity measures ( e.g. CP and Tucker decompositions [31]), the KBR has advantages in measuring the capacity of tensor space and unifying the traditional sparsity measures in case of 1-order and 2-order. Thus, it was proposed and applied to multispectral image denoising [33-35], and it obtained a great success in low-dose dynamic cerebral perfusion reconstruction [36].

Because Eq. (1) contains $L_0$-norm and low-rank terms, it is hard to optimize this problem. In practice, the KBR is relaxed as a log-sum form [33, 34] and Eq. (1) can be rewritten as

$$m(\mathcal{X}) = f(\mathcal{C}) + \alpha \prod_{n=1}^{N} f^*(X_{(n)}), \quad (2)$$

where

$$f(\mathcal{C}) = \sum_{i_1, i_2, \ldots, i_N}^{I_1, I_2, \ldots, I_N} \left( log(|c_{i_1, i_2, \ldots, i_N}| + \epsilon) - log(\epsilon) \right) / (-log(\epsilon))$$

$$f^*(X_{(n)}) = \sum_{q} \left( log(\sigma_q(X_{(n)}) + \epsilon) - log(\epsilon) \right) / (-log(\epsilon))$$

are two log-sum forms [33], $\epsilon$ is a small positive number and $\sigma_q(X_{(n)})$ defines the $q^{th}$ singular value of $X_{(n)}$. In this work, we employ this relaxation form to approximate the KBR measure.

### 2.B. Spectral CT Imaging Model

Considering the noise in projections, the conventional forward model for fan-beam CT scanning geometry can be discretized as a linear system

$$y = \mathcal{H}x + \eta, \quad (3)$$

where $x \in \mathcal{R}^{N_I}$ ($N_I = N_W \times N_H$) represents the vectorized 2D image, $y \in \mathcal{R}^J$ ($J = J_1 \times J_2$) stands for the vectorized projections, $J_1$ and $J_2$ are respectively the view and detector numbers and $\eta \in \mathcal{R}^J$ stands for projection noise. $\mathcal{H} \in \mathcal{R}^{J \times N_I}$ is the CT system matrix. Because $\mathcal{H}$ usually is too large to be saved in memory, Eq. (3) cannot be solved directly by matrix inverse techniques. Alternatively, it can be iteratively solved by minimizing the following optimization problem

$$\underset{x}{\operatorname{argmin}} \quad \frac{1}{2}\|y - \mathcal{H}x\|_2^2 \quad , \tag{4}$$

where $\|.\|_2$ represents the $L_2$ norm. Here, Eq. (4) can be minimized by the ART or SART methods [39]. To obtain a better solution, a regularization term of prior knowledge can be introduced and we have

$$\underset{x}{\operatorname{argmin}} \quad \frac{1}{2}\|y - \mathcal{H}x\|_2^2 + \lambda R(x) \quad . \tag{5}$$

Eq. (5) contains two terms, i.e., data fidelity term $\frac{1}{2}\|y - \mathcal{H}x\|_2^2$ and regularization term $R(x)$, and $\lambda > 0$ is a parameter to balance the data fidelity and regularization term.

For the spectral CT, because the emitted x-ray spectrum is divided into several narrow energy bins, the detectors can collect multiple projection datasets of the same imaging object with one scan, and each projection dataset can reconstruct one energy-dependent image. The model of fan-beam spectral CT reconstruction can be expressed as

$$\underset{\mathcal{X}}{\operatorname{argmin}} \quad \sum_{s=1}^{S} \frac{1}{2}\|y_s - \mathcal{H}x_s\|_2^2 + \lambda R(\mathcal{X}) \quad , \tag{6}$$

where $x_s$ is the vectorized image in $s^{th}$ ($s = 1,2,\dots,S$) energy channel, $y_s$ is the $s^{th}$ energy channel projection, and $\mathcal{X} \in \mathcal{R}^{N_W \times N_H \times S}$ is a 3rd-order tensor representing the set $\{x_s\}_{s=1}^{S}$.

The most important issue to reconstruct spectral CT image from its projections is to rationally extract prior structure knowledge and fully utilize such prior information to regularize the reconstruction. Due to the advantages of KBR regularizer, in this work, the KBR prior is incorporated into spectral CT reconstruction. That is, $R(\mathcal{X})$ in Eq. (6) is replaced by the aforementioned $m(\mathcal{X})$ defined in Eq.(2)

$$\underset{\mathcal{X}}{\operatorname{argmin}} \quad \sum_{s=1}^{S} \frac{1}{2}\|y_s - \mathcal{H}x_s\|_2^2 + \lambda m(\mathcal{X}). \tag{7}$$

*2.C. Non-local Similar Cubes Matching*

The noise in multi-energy projections can compromise the quality of the reconstructed image. Effectively implementing image reconstruction requires fully exploring prior knowledge. In this work, we mainly focus on the following three aspects. First, the human body usually only consist of two or three basis materials, i.e., soft tissue, bone and water in clinical applications. The number of basis materials is less than energy channels. This indicates the spectral images contain a large amount of spectral redundancy, and the images obtained from different channels are highly correlated [40, 41]. Second, as the multi-energy projection datasets are obtained from the same patient by different energy thresholds, images reconstructed across spectral dimension have different attenuation coefficients but share the same image structures. Third, small patches among different locations share similar structure information in a single channel image. It has been shown that such prior knowledge are very helpful for spectral CT reconstruction [7]. In a discrete image, one pixel only has 4 directly adjoin pixels. This corresponds to locally horizontal and vertical structures. Because a series of non-local similar cubes around the current cube are extracted to construct a 4D group, the image edges and structures along the horizontal and vertical directions can be well preserved in such a 4D group. Fig. 1 demonstrates the process of grouping. To explore non-local similarity inside an image, the patch length in spatial domain is usually very small (for examples, $4 \times 4$ and $6 \times 6$). This makes it difficult to accurately extract the intrinsic subspace bases of the spatial information for the HOSVD [42]. Meanwhile, we cannot afford the computational cost and memory load of larger patch length in the spatial horizontal and vertical model.

Therefore, in this work, we construct a 3rd-order low-rank cube for spectral CT reconstruction (see Fig. 1). For one given cube with size $r_w \times r_h \times r_s$ within the whole 3D images, we search $t$ similar non-local cubes in a given local window. Then, these extracted $t + 1$ small cubes used to formulate a new cube with the size of $(r_w r_h) \times r_s \times (t + 1)$, where $(r_w r_h) \times r_s$ is the matrix formation of cube and $t$ is the number of the non-local similar cubes. The formatted cube simultaneously explores the spatial local sparsity (mode-1), the non-local similarity among spectral-spatial cubes (mode-2) and spectral correlation (mode-3), which would be good for tensor recovery. The constructed cube also provides a unified interpretation for the matrix-based recovery model. Especially, $r_s = 1$ or $t = 0$, the constructed cube can be degenerated into a matrix by taking only non-local self-similarity or spectral correlation.

To better explore the similarities across spectral space, we set $r_s = S$ in the 3D low-rank cube construction. By traversing all the cubes across the spectral images with overlaps, we can build a set of 2D patches $\{\mathcal{G}_{ij} | 1 \leq i \leq N_W - r_w, 1 \leq j \leq N_H - r_h\} \subset \mathcal{R}^{(r_w r_h) \times S}$ to represent the spectral CT images, where each energy channel of a small patch is ordered lexicographically as a column vector. We can now reformulate all 2D patches as a group $\{\mathcal{X}_l\}_{l=1}^{L}$ with one index $l$, where $L = (N_W - r_w + 1) \times (N_H - r_h + 1)$ is the patch number. Furthermore, for a given current $\mathcal{X}_l$, we can find $t \geq 1$ similar patches within a non-local window. A low-rank cube can be constructed and denoted as $\mathcal{X}_l \in \mathcal{R}^{(r_w r_h) \times S \times (t+1)}$. Each cube can be considered as an extraction from the original 3rd-order tensor $\mathcal{X}$ with the operator $E_l$. Here, $\mathcal{X}_l$ can be further expressed as

$$\mathcal{X}_l = E_l \mathcal{X} \quad . \tag{8}$$

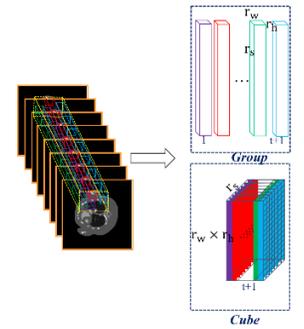

Figure 1. Illustration of the formation process for a group and cube.

*2.D. NLCTF Reconstruction Model*

Considering the non-local spatial similarity and correlation across spectral dimension in spectral images, we now construct a KBR-based non-local spectral CT reconstruction model based on Eq. (7),

$$\underset{\mathcal{X}}{\operatorname{argmin}} \sum_{s=1}^{S} \frac{1}{2} \|y_s - \mathcal{H}x_s\|_2^2 + \lambda \sum_{l=1}^{L} m(\mathcal{X}_l). \quad (9)$$

Substituting Eq. (8) into Eq.(9), we have

$$\underset{\mathcal{X}}{\operatorname{argmin}} \sum_{s=1}^{S} \frac{1}{2} \|y_s - \mathcal{H}x_s\|_2^2 + \lambda \sum_{l=1}^{L} m(E_l \mathcal{X}). \quad (10)$$

To optimize the problem Eq.(10), the split-Bregman method is employed. We split $\mathcal{X}$ and $E_l \mathcal{X}$ by introducing $L$ auxiliary cubes $\{\mathcal{T}_l\}_{l=1}^{L}$ instead of $\{E_l \mathcal{X}\}_{l=1}^{L}$, and Eq. (10) is rewritten as

$$\underset{\mathcal{X},\, \{\mathcal{T}_l\}_{l=1}^{L}}{\operatorname{argmin}} \frac{1}{2} \sum_{s=1}^{S} \|y_s - \mathcal{H}x_s\|_2^2 + \lambda \sum_{l=1}^{L} m(\mathcal{T}_l),$$
$$s.t.\, \mathcal{T}_l = E_l \mathcal{X} \quad (l = 1, \dots L). \quad (11)$$

Eq. (11) is a constrained optimization problem, which can be converted into an unconstrained version

$$\underset{\mathcal{X},\, \{\mathcal{T}_l, \mathcal{W}_l\}_{l=1}^{L}}{\operatorname{argmin}} \frac{1}{2} \sum_{s=1}^{S} \|y_s - \mathcal{H}x_s\|_2^2 + \lambda \sum_{l=1}^{L} m(\mathcal{T}_l)$$
$$+ \frac{\mu}{2} \sum_{l=1}^{L} \|\mathcal{T}_l - E_l \mathcal{X} - \mathcal{W}_l\|_F^2. \quad (12)$$

where the Frobenius norm of a tensor is used, $\mu$ is the coupling parameter and $\{\mathcal{W}_l\}_{l=1}^{L}$ represent $L$ error feedback cubes. Eq. (12) is equivalent to the following three sub-problems:

$$\begin{cases} \underset{\mathcal{X}}{\operatorname{argmin}} \left\{ \begin{array}{l} \frac{1}{2} \sum_{s=1}^{S} \|y_s - \mathcal{H}x_s\|_2^2 + \\ \frac{\mu}{2} \sum_{l=1}^{L} \|\mathcal{T}_l^{(k)} - E_l \mathcal{X} - \mathcal{W}_l^{(k)}\|_F^2 \end{array} \right\}, & (13a) \\ \underset{\{\mathcal{T}_l\}_{l=1}^{L}}{\min} \lambda \sum_{l=1}^{L} m(\mathcal{T}_l) + \frac{\mu}{2} \sum_{l=1}^{L} \|\mathcal{T}_l - E_l \mathcal{X}^{(k+1)} - \mathcal{W}_l^{(k)}\|_F^2, & (13b) \\ \underset{\{\mathcal{W}_l\}_{l=1}^{L}}{\operatorname{argmin}} \frac{\mu}{2} \sum_{l=1}^{L} \|\mathcal{T}_l^{(k+1)} - E_l \mathcal{X}^{(k+1)} - \mathcal{W}_l\|_F^2. & (13c) \end{cases}$$

where $k$ is the current iteration number and Eqs. (13a) – (13c) can be alternatingly solved.

Eq. (13a) can be solved by utilizing a gradient descent method. Its solution can be given as

$$\mathcal{X}_{n_w n_h S}^{(k+1)} = \mathcal{X}_{n_w n_h S}^{(k)} - \beta \left[ \mathcal{H}^T (\mathcal{H} x_s^{(k)} - y_s) \right]_{n_w n_h}$$
$$- \mu \left[ \sum_{l=1}^{L} E_l^{-1} (E_l \mathcal{X}^{(k)} - \mathcal{T}_l^{(k)} + \mathcal{W}_l^{(k)}) \right]_{n_w n_h S}. \quad (14)$$

where the symbols $[.]_{n_w n_h} (1 \leq n_w \leq N_W, 1 \leq n_h \leq N_H)$ and $[.]_{n_w n_h S}$ respectively indicate the $(n_w, n_h)^{th}$ and $(n_w, n_h, s)^{th}$ elements within a matrix and a tensor, $E_l^{-1}$ is the inverse operation of $E_l$, and $\beta \in (0,2)$ is a relaxation factor, which is set as 0.03 in our experiments [7]. Eq. (13c) can be easily solved by using the steepest descent method

$$\mathcal{W}_l^{(k+1)} = \mathcal{W}_l^{(k)} - \wp_l (\mathcal{T}_l^{(k+1)} - E_l \mathcal{X}^{(k+1)}), \forall l = 1, \dots L, \quad (15)$$

where $\wp_l$ is the length of step size and it is set as 1.0 in this work. Now, the challenge is to solve the problem (13b). Substituting Eq. (2) into Eq. (13b), we have

$$\underset{\{\mathcal{T}_l\}_{l=1}^{L}}{\operatorname{argmin}} \sum_{l=1}^{L} \left( f(\mathcal{C}_l) + \alpha \prod_{n=1}^{3} f^* (\mathcal{T}_{l_{(n)}}) \right)$$
$$+ \frac{\delta}{2} \sum_{l=1}^{L} \|\mathcal{T}_l - E_l \mathcal{X}^{(k+1)} - \mathcal{W}_l^{(k)}\|_F^2, \quad (16)$$

where $\mathcal{T}_{l_{(n)}}$ represents the unfolding matrix with the mode-$n$ of the tensor $\mathcal{T}_l$. $\delta = \mu/\lambda$ and $N = 3$. Eq. (16) can be divided into $L$ independent sub-problems

$$\underset{\mathcal{T}_l}{\operatorname{argmin}} f(\mathcal{C}_l) + \alpha \prod_{n=1}^{3} f^* (\mathcal{T}_{l_{(n)}})$$
$$+ \frac{\delta}{2} \|\mathcal{T}_l - E_l \mathcal{X}^{(k+1)} - \mathcal{W}_l^{(k)}\|_F^2. \quad (17)$$

To minimize an objective function similar to Eq. (17), the ADMM was proposed in [33]. Compared with ADMM method, the split-Bregman can simplify the minimization step by decoupling the variables coupled by the constraint matrix $I$ in the Eq. (17)[43, 44]. Besides, because the regularization function $f(\mathcal{C}_l) + \alpha \prod_{n=1}^{3} f^* (\mathcal{T}_{l_{(n)}})$ is a nonconvex function, the split-Bregman outperforms the ADMM to optimize Eq. (17) [43-45]. Finally, because of the faster convergence and easier implementation than ADMM, the split-Bregman method is increasingly becoming a method of choice for solving sparsity recovery problems [46-48]. Here, we adopt the split-Bregman method instead of ADMM and then follow similar steps in [33] to obtain the final solution. First, we need to introduce $3L$ auxiliary cubes $\{\mathcal{M}_{l_n}\}_{n=1}^{3}$. Eq. (17) can be written as

$$\underset{\{\mathcal{M}_{l_n}, \mathcal{Q}_{l_n}\}_{n=1}^{3}, \mathcal{C}_l}{\operatorname{argmin}} f(\mathcal{C}_l) + \alpha \prod_{n=1}^{3} f^* (\mathbf{M}_{l_{n(n)}})$$
$$+ \frac{\delta}{2} \|\mathcal{C}_l \times_1 \mathcal{Q}_{l_1} \times_2 \mathcal{Q}_{l_2} \times_3 \mathcal{Q}_{l_3} - E_l \mathcal{X}^{(k+1)} - \mathcal{W}_l^{(k)}\|_F^2$$
$$s.t., \mathcal{C}_l \times_1 \mathcal{Q}_{l_1} \times_2 \mathcal{Q}_{l_2} \times_3 \mathcal{Q}_{l_3} = \mathcal{M}_{l_n} (n = 1,2,3),$$
$$\mathcal{Q}_{l_n}^T \mathcal{Q}_{l_n} = I \ (n = 1,2,3), \quad (18)$$

where the factor matrices $\{\mathcal{Q}_{l_n}\}_{n=1}^{3}$ denote orthogonal in columns and $\mathbf{M}_{l_{n(n)}}$ represent the unfolding matrix along mode-$n$ of the cube $\mathcal{M}_{l_n}$. Eq. (18) is a constrained problem which can be converted into an unconstrained one

$$\underset{\mathcal{C}_l, \{\mathcal{M}_{l_n}, \mathcal{Z}_{l_n}, \mathcal{Q}_{l_n}\}_{n=1}^{3}}{\operatorname{argmin}} f(\mathcal{C}_l) + \alpha \prod_{n=1}^{3} f^* (\mathbf{M}_{l_{n(n)}})$$
$$+ \frac{\delta}{2} \|\mathcal{C}_l \times_1 \mathcal{Q}_{l_1} \times_2 \mathcal{Q}_{l_2} \times_3 \mathcal{Q}_{l_3} - E_l \mathcal{X}^{(k+1)} - \mathcal{W}_l^{(k)}\|_F^2$$
$$+ \frac{\theta}{2} \sum_{n=1}^{3} \|\mathcal{C}_l \times_1 \mathcal{Q}_{l_1} \times_2 \mathcal{Q}_{l_2} \times_3 \mathcal{Q}_{l_3} - \mathcal{M}_{l_n} + \mathcal{Z}_{l_n}\|_F^2, \quad (19)$$

where $\{\mathcal{Z}_{l_n}\}_{n=1}^{3}$ represent errors feedback cube, $\theta$ is a positive parameter and $\{\mathcal{Q}_{l_n}\}_{n=1}^{3}$ satisfy $\mathcal{Q}_{l_n}^T \mathcal{Q}_{l_n} = I$. Now Eq. (19) can be updated by solving the following sub-problem:

*i) $\mathcal{C}_l$ sub-problem:* With the other parameters fixed, $\mathcal{C}_l$ can be updated by solving the following minimization problem:

$$\min_{\mathcal{C}_l} \gamma f(\mathcal{C}_l) + \frac{1}{2} \left\| \mathcal{C}_l \times_1 \mathbf{Q}_{l_1}^{(k)} \times_2 \mathbf{Q}_{l_2}^{(k)} \times_3 \mathbf{Q}_{l_3}^{(k)} - \mathcal{B}_l^{(k)} \right\|_F^2. \quad (20)$$

where $\gamma = 1/(\delta + 3\theta)$ and $\mathcal{B}_l^{(k)} = \left( \delta(E_l \mathcal{X}^{(k+1)} + \mathcal{W}_l^{(k)}) + \theta(\sum_{n=1}^3 (\mathcal{M}_{l_n}^{(k)} - \mathcal{Z}_{l_n}^{(k)})) \right) / (\delta + 3\theta)$. Based the results in [33], Eq. (20) can be converted to

$$\min_{\mathcal{C}_l} \gamma f(\mathcal{C}_l) + \frac{1}{2} \left\| \mathcal{C}_l - \mathcal{D}_l^{(k)} \right\|_F^2, \quad (21)$$

where

$$\mathcal{D}_l^{(k)} = \mathcal{B}_l^{(k)} \times_1 \left( \mathbf{Q}_{l_1}^{(k)} \right)^T \times_2 \left( \mathbf{Q}_{l_2}^{(k)} \right)^T \times_3 \left( \mathbf{Q}_{l_3}^{(k)} \right)^T.$$

Eq. (21) has a closed-form solution [49]

$$\mathcal{C}_l^{(k+1)} = D_{\gamma,\epsilon}(\mathcal{D}_l^{(k)}), \quad (22)$$

where $D_{\gamma,\epsilon}(\cdot)$ denotes the hard-thresholding operation, which has the following form

$$D_{\gamma,\epsilon}(x) = \begin{cases} 0 & \text{if } |x| \le 2\sqrt{c_1\gamma} - \epsilon \\ sign\left( \frac{c_2(x) + c_3(x)}{2} \right) & \text{if } |x| > 2\sqrt{c_1\gamma} - \epsilon \end{cases}, \quad (23)$$

where $c_1 = (-1)/log(\epsilon)$, $c_2(x) = |x| - \epsilon$, $c_3(x) = \sqrt{(|x| + \epsilon)^2 - 4c_1\gamma}$ and $sign$ represents the sign function.

*ii)* $\{\mathbf{Q}_{l_n}\}_{n=1}^3$ *sub-problem:* with respect to $\mathbf{Q}_{l_1}$, we fix the $\mathbf{Q}_{l_2}^{(k)}$, $\mathbf{Q}_{l_3}^{(k)}$ and others parameters. $\mathbf{Q}_{l_1}$ can be updated by minimizing the following problem

$$\min_{\mathbf{Q}_{l_1}} \frac{1}{2} \left\| \mathcal{C}_l^{(k+1)} \times_1 \mathbf{Q}_{l_1} \times_2 \mathbf{Q}_{l_2}^{(k)} \times_3 \mathbf{Q}_{l_3}^{(k)} - \mathcal{B}_l^{(k)} \right\|_F^2$$

$$s.t. \left( \mathbf{Q}_{l_1} \right)^T \mathbf{Q}_{l_1} = \mathbf{I}. \quad (24)$$

According the work in [36], Eq. (24) is equivalent to

$$\max_{(\mathbf{Q}_{l_1})^T \mathbf{Q}_{l_1} = \mathbf{I}} \langle \mathcal{L}_{l_1}, \mathbf{Q}_{l_1} \rangle, \quad (25)$$

where $\mathcal{L}_{l_1} = \mathbf{B}_{l(1)}^{(k)} \left( \mathbf{Q}_{l_2}^{(k)} \otimes \mathbf{Q}_{l_3}^{(k)} \right) \left( \mathbf{C}_{l(1)}^{(k+1)} \right)^T$ and $\mathbf{B}_{l(1)}^{(k)}$ represent the unfolding matrix of $\mathcal{B}_l^{(k)}$ along mode-1. Then, $\mathbf{Q}_{l_1}$ can be updated by [33]

$$\mathbf{Q}_{l_1}^{(k+1)} = \mathbf{G}_{l_1} \left( \mathbf{V}_{l_1} \right)^T, \quad (26)$$

where $\mathbf{G}_{l_1} \mathbf{\Theta}_{l_1} \left( \mathbf{V}_{l_1} \right)^T$ represents the SVD decomposition of $\mathcal{L}_{l_1}$. Similarly, $\mathbf{Q}_{l_2}$ and $\mathbf{Q}_{l_3}$ can be updated by minimizing

$$\begin{cases} \min_{(\mathbf{Q}_{l_2})^T \mathbf{Q}_{l_2} = \mathbf{I}} \frac{1}{2} \left\| \mathcal{C}_l^{(k+1)} \times_1 \mathbf{Q}_{l_1}^{(k+1)} \times_2 \mathbf{Q}_{l_2} \times_3 \mathbf{Q}_{l_3}^{(k)} - \mathcal{B}_l^{(k)} \right\|_F^2 \\ \min_{(\mathbf{Q}_{l_3})^T \mathbf{Q}_{l_3} = \mathbf{I}} \frac{1}{2} \left\| \mathcal{C}_l^{(k+1)} \times_1 \mathbf{Q}_{l_1}^{(k+1)} \times_2 \mathbf{Q}_{l_2}^{(k+1)} \times_3 \mathbf{Q}_{l_3} - \mathcal{B}_l^{(k)} \right\|_F^2 \end{cases}. \quad (27)$$

*iii)* $\{\mathcal{M}_{l_n}\}_{n=1}^3$ *sub-problem:* To update $\mathcal{M}_{l_1}$, we fix $\mathcal{M}_{l_2}^{(k)}$, $\mathcal{M}_{l_3}^{(k)}$ another parameters. The update of $\mathcal{M}_{l_1}$ can be obtained by minimizing

$$b_{l_1} f^*(\mathbf{M}_{l_{1(1)}}) + \frac{1}{2} \left\| \mathcal{M}_{l_1} - \mathcal{Q}_l^{(k+1)} - \mathcal{Z}_{l_1}^{(k)} \right\|_F^2, \quad (28)$$

where

$$\mathcal{Q}_l^{(k+1)} = \mathcal{C}_l^{(k+1)} \times_1 \mathbf{Q}_{l_1}^{(k+1)} \times_2 \mathbf{Q}_{l_2}^{(k+1)} \times_3 \mathbf{Q}_{l_3}^{(k+1)}. \quad (29)$$

The $b_{l_e}$ is defined as $b_{l_e} = \frac{\alpha}{\theta} \prod_{e \ne n} f^*(\mathbf{M}_{l_{n(n)}})$ and $e = 1,2,3$, i.e., $b_{l_1}$ in Eq. (29) can be expressed as

$$b_{l_1} = \frac{\alpha}{\theta} \prod_{n=2,3} f^*(\mathbf{M}_{l_{n(n)}}).$$

According to Theorem 1 in [50] and the work in [33], Eq. (29) has the following closed-form solution:

$$\mathcal{M}_{l_1}^{(k+1)} = \text{fold}_1 \left( \mathbf{\Psi}_{l_1} \mathbf{\Sigma}_{d_{l_1}} \mathbf{\psi}_{l_1}^T \right), \quad (30)$$

where $\mathbf{\Sigma}_{d_{l_1}} = diag \left( D_{d_{l_1},\epsilon}(\sigma_1), D_{d_{l_1},\epsilon}(\sigma_2), \dots, D_{d_{l_1},\epsilon}(\sigma_M) \right)$ and $\mathbf{\Psi}_{l_1} diag(\sigma_1, \sigma_2, \dots, \sigma_M) \mathbf{\psi}_{l_1}^T$ is the SVD decomposition of $\text{unfold}_1(\mathcal{Q}_l^{(k+1)} + \mathcal{Z}_{l_1}^{(k)})$. $\mathcal{M}_{l_2}$ and $\mathcal{M}_{l_3}$ can be updated in a similar way.

*iv)* $\{\mathcal{Z}_{l_n}\}_{n=1}^3$ *sub-problem:* From Eqs. (19), (28) and (29), $\mathcal{Z}_{l_n}$ can be updated as

$$\mathcal{Z}_{l_n}^{(k+1)} = \mathcal{Z}_{l_n}^{(k)} - (\mathcal{M}_{l_n}^{(k+1)} - \mathcal{Q}_l^{(k+1)}). \quad (31)$$

All the aforementioned steps in the proposed NLCTF method can be summarized in Algorithm I. Note that all cubes are constructed by the normalized $\mathcal{X}^{(k+1)}$ not the original $\mathcal{X}^{(k+1)}$. Thus, it is necessary to denormalize the updated $\mathcal{T}_l^{(k+1)} (l = 1, \dots L)$. For the formulation of a low-rank cube, $r_w$, $r_h$ and $t$ are set as 6, 6 and 50, respectively. The parameter $\delta$ depends on $\tau$ and $\delta = c\tau^{-1}$, where c is set as a constant $10^{-3}$. The size of the search window is set as $80 \times 80$ in this work.

---

**Algorithm I: NLCTF**

**Input**: $\{y_s\}_{s=1}^S$, $\alpha$, $\tau$, $\theta$, $\mu$ and other parameters;
1: Initialization: $\{\mathcal{X}^{(0)}\} \leftarrow \mathbf{0}$; $\{\mathcal{X}_l^{(0)}, \mathcal{W}_l^{(0)}\} \leftarrow \mathbf{0}$, initializing $\mathbf{Q}_{l_n}^{(0)}$, $\mathcal{C}_l^{(0)}$ by high-order SVD of $\mathcal{X}_l^{(0)}$; $\{\mathcal{M}_{l_n}^{(0)}, \mathcal{Z}_{l_n}^{(0)}, \mathcal{T}_l^{(0)}\} \leftarrow \mathcal{X}_l^{(0)}, \forall n = 1,2,3$ and $l = 1, \dots L$; $k = 0$;
2: **While** not convergence **do**
3:   Updating $\mathcal{X}^{(k+1)}$ using Eq. (14);
4:   Constructing all cubes $\mathcal{X}_l^{(k+1)}(l = 1, \dots L)$ using normalized $\mathcal{X}^{(k+1)}$ by Eq. (8);
6:   for $l = 1:L$
7:     Updating $\mathcal{C}_l^{(k+1)}$ using Eq. (22);
8:     Updating $\mathbf{Q}_{l_n}^{(k+1)}(n = 1,2,3)$ using Eq. (26);
9:     Updating $\mathcal{M}_{l_n}^{(k+1)}(n = 1,2,3)$ using Eq. (30);
10:    Updating $\mathcal{Z}_{l_n}^{(k+1)}(n = 1,2,3)$ using Eq. (31);
11:    Updating the denormalized
      $\mathcal{T}_l^{(k+1)} = \mathcal{C}_l^{(k+1)} \times_1 \mathbf{Q}_{l_1}^{(k+1)} \times_2 \mathbf{Q}_{l_2}^{(k+1)} \times_3 \mathbf{Q}_{l_3}^{(k+1)}$;
12:    Updating $\mathcal{W}_l^{(k+1)}$ using Eq. (15);
13:   End for
14:   $k = k + 1$;
15: **End while**
**Output**: $\mathcal{X}$

### 2.E. Comparison algorithms

To evaluate the performance of our proposed NLCTF algorithm, the SART, total variation minimization (TV)[8], total variation and low rank (TV+LR) [10], image gradient $L_0$-norm and tensor dictionary learning ($L_0$TDL) [26], as well as the spatial-spectral cube matching frame (SSCMF) [7] algorithms are chosen and implemented for comparison. It should be emphasized that all hyper-parameters in the TV, TV+LR, $L_0$TDL and SSCMF methods are empirically optimized in our experiments.

### III. EXPERIMENTS AND RESULTS

In this section, projections from both numerically simulated mouse thorax phantom and real mouse with injected gold nanoparticles (GNP) are employed to validate and evaluate the developed NLCTF algorithm. In numerical simulations, we mainly demonstrate the performance of our proposed method in terms of reconstructed image quality, material decomposition accuracy, algorithm convergence, and computational cost. To quantitatively evaluate the image quality, the root mean square error (RMSE), peak-signal-to-noise ratio (PSNR), feature similarity (FSIM) [51] and structural similarity (SSIM) are employed. The results from preclinical real mouse data also confirm the outperformance of the proposed method in recovering finer structures and preserving image edges with reduced noise.

### 3.A. Numerical Simulation Study

A numerical mouse thorax phantom injected with a 1.2% iodine contrast agent is used (see Fig.2 (a)), and a polychromatic 50KVp x-ray source is assumed whose normalized spectrum is given in Fig. 2(b). The spectrum is divided into eight energy bins: [16, 22) keV, [22, 25) keV, [25, 28) keV, [28, 31) keV, [31, 34) keV, [34, 37) keV, [37, 41) keV, and [41, 50) keV. The PCD includes 512 elements, and the length of each element is 0.1mm. The distances from x-ray source to PCD and object are set as 180mm and 132mm, generating a field of view (FOV) of 37.2 mm in diameter. 640 projections are acquired over a full scan. Poisson noise is superimposed, and the photon number emitting from the x-ray source for each x-ray path is set as $2\times10^4$. All the energy bin images of 512×512 matrixes from different methods are reconstructed after 50 iterations.

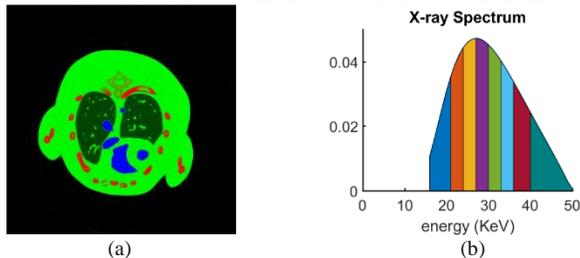

(a)          (b)
Figure 2. Numerical simulation setup. (a) is the mouse thorax phantom, where green, red and blue stand for water, bone and iodine, respectively. (b) is the normalized x-ray source spectrum.

The parameter selection is challenging for the proposed NLCTF method. There are four key parameters, i.e., $\alpha$, $\delta$, $\theta$, and $\mu$. To make it clear, these parameters are summarized in Table I. Other parameters in the competing algorithms are also optimized, and the best results are selected for comparison and analysis.

Table I. NLCTF parameters.

| Methods | Photon No. | $\alpha$ | $\tau$ | $\theta$ | $\mu$ |
|---|---|---|---|---|---|
| Numerical Simulation | $2\times10^4$ | 10 | 0.050 | 250 | 0.5 |
| Preclinical Application | - | 10 | 0.075 | 250 | 0.5 |

#### 1). Reconstruction Results

Fig.3 shows three representative energy channels (1st, 4th and 8th) of the reconstructed images, where the reference images are reconstructed from noise-free projections by the SART. It can be seen that the images reconstructed by the SSCMF and $L_0$TDL methods have much finer structures and details compared with those are reconstructed by the TV+LR method followed by the TV algorithm. Without any prior knowledge in the mathematical model, there are always strongest noise and image artifacts in the SART results.

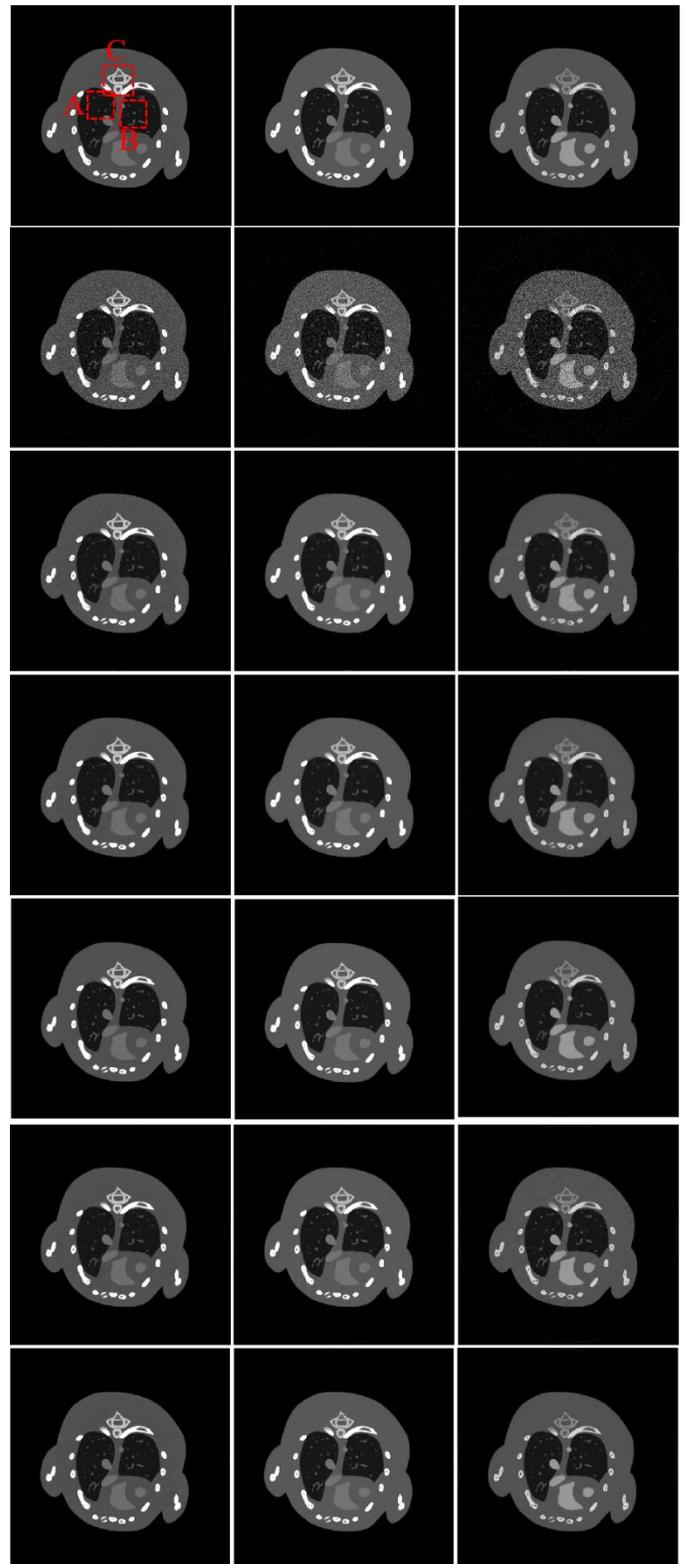

Figure 3. Three representative energy bin reconstruction results. The 1st row are reference images, and the 2nd to 7th row images are reconstructed by using the SART, TV, TV+LR, $L_0$TDL, SSCMF and NLCTF methods. The 1st to 3rd columns correspond to the 1st, 4th and 8th channels, and their display windows are [0, 3], [0, 1.2] and [0, 0.8] cm$^{-1}$, respectively.

It can be seen that the reconstructed image quality from the TV and TV+LR methods are improved (3rd and 4th rows) with prior knowledge. Because both the TV and TV+LR models contain TV regularization, their results have blocky artifacts and blurry image edges. The $L_0$TDL can provide better images than the TV and TV+LR methods with significantly reduced blocky artifacts. Compared with the $L_0$TDL method, the SSCMF can retain much finer structures. However, some finer structures and details are still lost in the SSCMF results. In comparison, the NLCTF algorithm achieves a great success in capturing smaller image structures and details.

To clearly compare the reconstruction performance of all algorithms, two lung ROIs ("A" and "B") and one bony ROI ("C") are extracted from Fig. 3 and magnified in Fig. 4. From Fig. 4(a), we can observe that the finer details indicated by the arrows "1" and "2" can still be seen in the NLCTF results, but these structures are lost by other competitors. The image structure indicated by the arrow "3" has disappeared in TV, TV+LR results and blurred from SSCMF. However, it is persevered well in the $L_0$TDL and NLCTF methods even if it is slightly blurred from the $L_0$TDL, especially in high energy bins. The feature indicated by the arrow "4" cannot be seen in the TV and TV+LR results. Although this feature can be found in low energy bins, it is invisible in high energy bin results of the $L_0$TDL and SSCMF. However, it can always be seen in the NLCTF results. Fig. 4 (b) shows another lung ROI "B". From Fig. 4 (b), we can see that the image structures indicated by arrows "5" and "6" are lost by other competing algorithms, and these structures can still be faithfully reconstructed by the NLCTF algorithm. Fig. 4 (c) shows a magnified bony ROI "C", where the thoracic vertebra bones are separated by low-density tissues. It can be seen that the SART results contain severe noise, especially in the 8th channel where the signal-to-noise ratio (SNR) is too low to distinguish the thoracic vertebra bony structures and soft tissue. The SSCMF results preserve more bony structures than the TV and TV+LR techniques. However, the image edges surrounding the bony region are still not clear. Generally speaking, both the $L_0$TDL and NLCTF can offer high quality images with sharp image edges and reduced image noise. However, the image edge indicated by the arrow "8" is still slightly blurred.

Table II shows the quantitative evaluation results of the reconstructed images from all energy bins. It shows that the proposed NLCTF can always obtain the smallest RMSEs for all energy bins. From table II, we can see that the TV+LR method has slightly smaller RMSEs than the TV. Compared with the TV and TV+LR methods, the $L_0$TDL and SSCMF have better reconstruction performance. Specifically, the RMSE values of $L_0$TDL are greater than those obtained by the SSCMF method in lower energy bins (1$^{st}$, 2$^{nd}$), and the $L_0$TDL has smaller RMSE values than those achieved by the SSCMF in higher energy bins (3$^{rd}$-8$^{th}$). In terms of PSNR, similar conclusions can be made. The SSIM and FSIM measure the similarity between the reconstructed images and references, which are recently employed to compare reconstructed CT image quality [52]. Here, the dynamic range of all channel images are scaled to [0 255]. The closer to 1.0 the SSIM and FSIM values are, the better the reconstructed image quality is. In Table II, the NLCTF results obtain the greatest SSIM and FISM values for all channels all the time. In terms of these two indexes, the $L_0$TDL and SSCMF outperform the TV and TV+LR methods. In general, the NLCTF method has the higher image quality in terms of quantitative assessment.

Table II. Quantitative evaluation results of the reconstructed images and material decomposition

| Index | Method | Reconstructed images (Channel Number) | | | | | | | | Material decomposition | | |
|---|---|---|---|---|---|---|---|---|---|---|---|---|
| | | 1$^{st}$ | 2$^{nd}$ | 3$^{rd}$ | 4$^{th}$ | 5$^{th}$ | 6$^{th}$ | 7$^{th}$ | 8$^{th}$ | Bone | Soft tissue | Iodine |
| RMSE ($10^{-2}$) | SART | 11.88 | 8.50 | 6.85 | 6.23 | 6.07 | 6.35 | 6.00 | 5.72 | 1.27 | 6.77 | 4.96 |
| | TVM | 3.80 | 2.57 | 1.98 | 1.73 | 1.55 | 1.53 | 1.37 | 1.21 | 0.70 | 2.61 | 1.95 |
| | TV+LR | 3.69 | 2.34 | 1.73 | 1.40 | 1.22 | 1.34 | 1.14 | 0.98 | 0.85 | 2.67 | 1.87 |
| | $L_0$TDL | 3.18 | 1.90 | 1.32 | 1.00 | 0.81 | 0.81 | 0.70 | 0.63 | 0.75 | 1.75 | **0.99** |
| | SSCMF | 2.68 | 1.62 | 1.54 | 1.03 | 1.03 | 0.88 | 0.78 | 0.73 | 0.50 | 1.90 | 1.19 |
| | NLCTF | **2.27** | **1.36** | **0.99** | **0.77** | **0.65** | **0.71** | **0.63** | **0.57** | **0.47** | **1.62** | **0.99** |
| PSNR | SART | 18.50 | 21.40 | 23.28 | 24.11 | 24.33 | 23.94 | 24.44 | 24.86 | 37.92 | 23.39 | 26.08 |
| | TVM | 28.40 | 31.81 | 34.08 | 35.26 | 36.17 | 36.31 | 37.23 | 38.34 | 43.11 | 31.68 | 34.20 |
| | TV+LR | 28.67 | 32.60 | 35.25 | 37.09 | 38.29 | 37.43 | 38.86 | 40.13 | 41.40 | 31.47 | 34.59 |
| | $L_0$TDL | 29.95 | 34.42 | 37.58 | 40.01 | 41.82 | 41.81 | 43.06 | 44.08 | 42.52 | 35.12 | **40.06** |
| | SSCMF | 31.44 | 35.83 | 38.53 | 39.74 | 41.10 | 41.14 | 42.11 | 42.76 | 46.06 | 34.44 | 38.50 |
| | NLCTF | **32.89** | **37.33** | **40.05** | **42.26** | **43.79** | **42.92** | **43.97** | **44.84** | **46.54** | **35.83** | **40.06** |
| SSIM | SART | 0.9125 | 0.8838 | 0.8647 | 0.8273 | 0.7737 | 0.7251 | 0.6801 | 0.6291 | 0.9118 | 0.6176 | 0.7596 |
| | TVM | 0.9901 | 0.9876 | 0.9835 | 0.9746 | 0.9595 | 0.9407 | 0.9220 | 0.8918 | 0.9976 | 0.8923 | 0.9470 |
| | TV+LR | 0.9909 | 0.9877 | 0.9843 | 0.9759 | 0.9684 | 0.9583 | 0.9318 | 0.9152 | 0.9949 | 0.8654 | 0.9544 |
| | $L_0$TDL | 0.9932 | 0.9958 | 0.9942 | 0.9905 | 0.9851 | 0.9744 | 0.9630 | 0.9610 | 0.9978 | 0.9380 | 0.9827 |
| | SSCMF | 0.9915 | 0.9956 | 0.9962 | 0.9949 | 0.9930 | 0.9911 | 0.9911 | 0.9865 | 0.9984 | 0.9687 | 0.9661 |
| | NLCTF | **0.9937** | **0.9982** | **0.9980** | **0.9977** | **0.9971** | **0.9956** | **0.9946** | **0.9925** | **0.9988** | **0.9689** | **0.9859** |
| FSIM | SART | 0.8822 | 0.8345 | 0.8223 | 0.8037 | 0.7797 | 0.7567 | 0.7567 | 0.7177 | 0.9899 | 0.8098 | 0.9084 |
| | TVM | 0.9780 | 0.9730 | 0.9713 | 0.9647 | 0.9497 | 0.9326 | 0.9326 | 0.9133 | 0.9985 | 0.9699 | 0.9578 |
| | TV+LR | 0.9870 | 0.9836 | 0.9815 | 0.9722 | 0.9564 | 0.9365 | 0.9365 | 0.9238 | 0.9981 | 0.9760 | 0.9581 |
| | $L_0$TDL | 0.9861 | 0.9924 | 0.9912 | 0.9876 | 0.9852 | 0.9823 | 0.9789 | 0.9715 | 0.9989 | **0.9878** | 0.9620 |
| | SSCMF | 0.9871 | 0.9909 | 0.9885 | 0.9855 | 0.9824 | 0.9794 | 0.9761 | 0.9711 | 0.9988 | 0.9755 | 0.9604 |
| | NLCTF | **0.9884** | **0.9934** | **0.9915** | **0.9897** | **0.9877** | **0.9832** | **0.9801** | **0.9748** | **0.9991** | 0.9767 | **0.9637** |

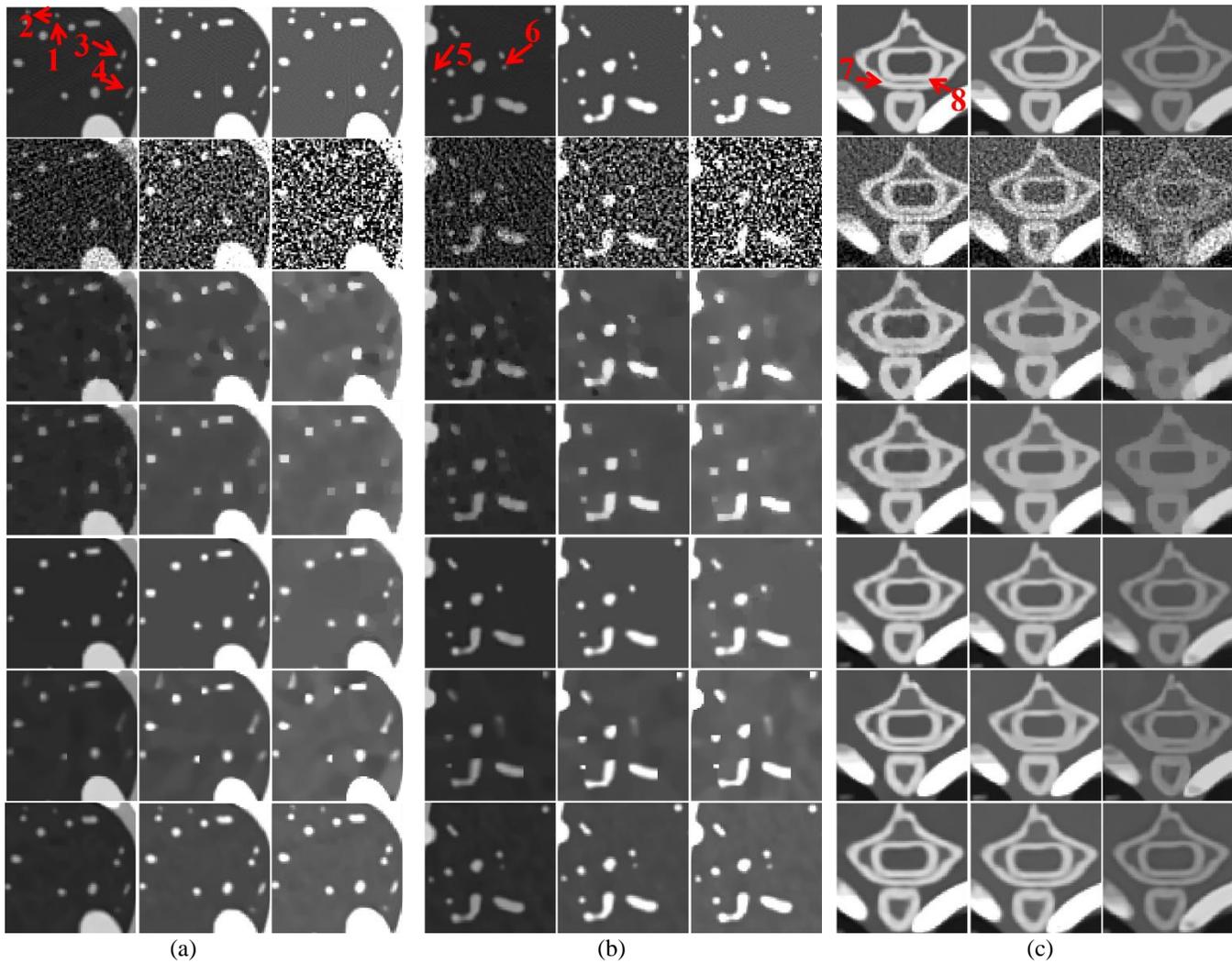

Figure 4. Magnified ROIs. (a), (b) and (c) correspond to the ROI A, B and C in Fig. 3. The display windows of both (a) and (b) are [0, 1.6], [0, 0.4] and [0, 0.2] cm$^{-1}$. The display window of (c) is the same as Fig. 3.

*2). Material Decomposition*

To evaluate all the algorithms for material decomposition, all the reconstructed spectral CT images are decomposed into three basis materials (soft tissue, iodine contrast agent and bone) utilizing a post-processing method [25]. Fig. 5 shows the three decomposed basis materials and the corresponding color rendering images. The first column of Fig. 5 shows the bone component. It can be seen that many soft tissue and iodine contrast agent pixels are wrongly introduced by the SART. Compared with the L$_0$TDL and SSCMF results, more pixels of the iodine from the TV and TV+LR are also wrongly classified as bone structure. However, there are still some pixels of iodine contrast agent that are wrongly classified as bone in the L$_0$TDL and SSCMF. In contrast, the NLCTF result has a clear bone map. Regarding the soft tissue component (2$^{nd}$ column), two ROIs indicated by "D" and "E" are extracted to evaluate the performance of all algorithms. From the extracted ROI "D", it can be seen that the image edges indicated by arrows are blurred in the TV, TV+LR and SSCMF results. The resolution of the image edges is significantly improved in the L$_0$TDL results. However, the image structures are still slightly blurred when they are compared with those obtained from the NLCTF reconstructions. In terms of ROI "E", finer structures indicated by arrows are lost in the competitors, and they are well preserved in the proposed NLCTF results. As for the iodine contrast agent results (3$^{rd}$ column in Fig. 5), the bony structures have an impact on the accuracy of the iodine contrast agent component, especially on the results from the SART. The accuracy of the iodine contrast agent is significantly improved in the TV and TV+LR results. However, some bony pixels are still wrongly decomposed. Compared with the results of the SSCMF, the L$_0$TDL and NLCTF provide much clear maps.

To further quantitatively evaluate the accuracy of material decomposition for all the reconstruction algorithms, the RMSE, SSIM, PSNR and FSIM values of three decomposed basis materials are also listed in table II, where the references are obtained from the SART results with noise-free projections. Table II demonstrates the NLCTF can obtain the smallest RMSE values for three different materials, even if the L$_0$TDL can obtain the same RMSE value for the iodine contrast agent. In terms of PSNR and SSIM, we can obtain similar conclusions. As for the FSIM index, the higher values of bone and iodine contrast agent can be obtained by the NLCTF method. The L$_0$TDL method can obtain greater FSIM value than other algorithms. However, the finer image details and small structures are lost in the L$_0$TDL results.

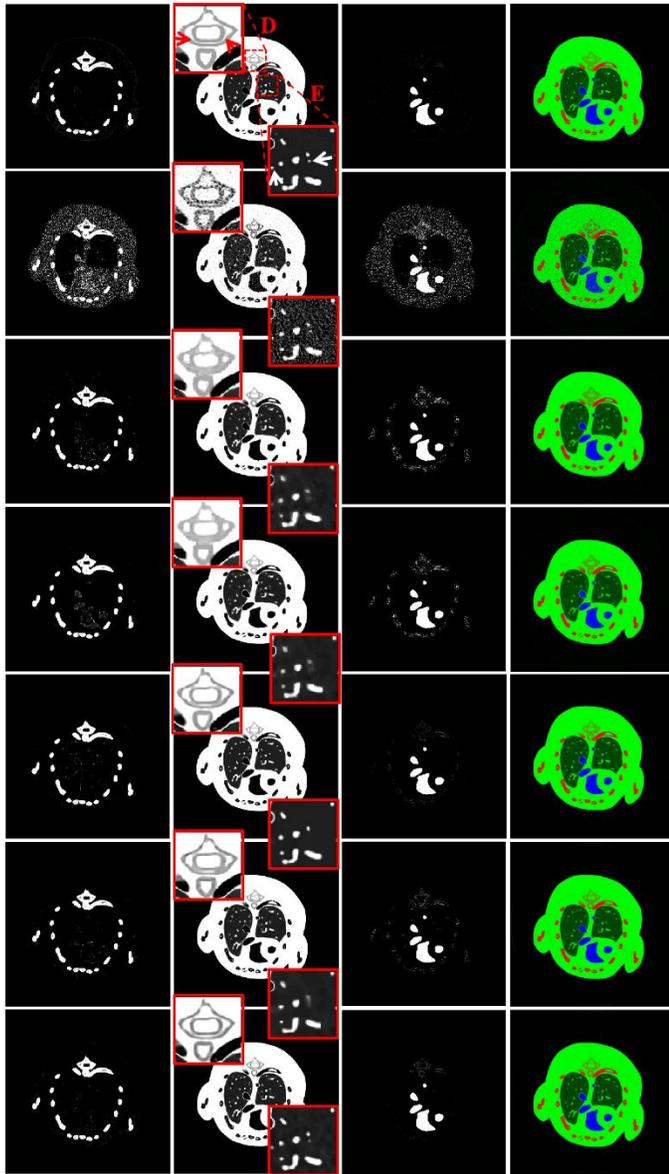

Figure 5. Three basis materials decomposed from the reconstructed results in Fig. 3. The 1$^{st}$ to 3$^{rd}$ columns represent bone, soft tissue and iodine, and their display windows are [0, 0.05], [0.2, 0.8] and [0, 0.25] cm$^{-1}$. The 4$^{th}$ column images are color rendering, where red, green and blue represent bone, soft tissue and iodine. The 1$^{st}$ row images are the references and 2$^{nd}$ to 7$^{th}$ rows are the decomposed results using the SART, TV, TV+LR, L$_0$TDL, SSCMF and NLCTF methods for reconstruction, respectively.

### 3). Parameters analysis

The parameters of NLCTF mainly comes from two parts: model regularization parameters ($\alpha$, $\mu$, $\delta$, $\theta$) and cube formulation parameters (patch size $r \times r$, similar patch number $t$). To investigate the effect of each parameter on the final reconstruction, the NLCTF results with respect to different parameter settings are compared. Here, the RMSE and SSIM are computed after 50 iterations for analysis. Fig. 6 shows the final RMSE and SSIM values of the NLCTF method with respect to different parameters, and each subplot represents the RMSE or SSIM values with respect to one varying parameter with other parameters are fixed.

It is observed from Fig. 6 that the parameters $\alpha$, $\mu$, and $\delta$ play an important role in controlling the reconstructed image quality. Specifically, an appropriate $\alpha$ can reduce RMSE value with greater SSIM value, while one smaller or greater $\alpha$ can increase the RMSE and reduce the SSIM. Regarding $\mu$ and $\delta$, similar conclusions can be made. According to the Fig. 6 (g) and (h), it can be observed that the parameter $\theta$ has a small impact on the RMSE values. However, it can improve the SSIM by selecting an appropriate value. From Fig. 6 (i) and (j), we can see that, when the number of similar patches is 50, we can obtain relatively optimized reconstructed results. For the optimal patch size, Fig. 6 (k) and (l) show the $6 \times 6$ can always obtain the smallest RMSE values with greatest SSIM values in all energy channels.

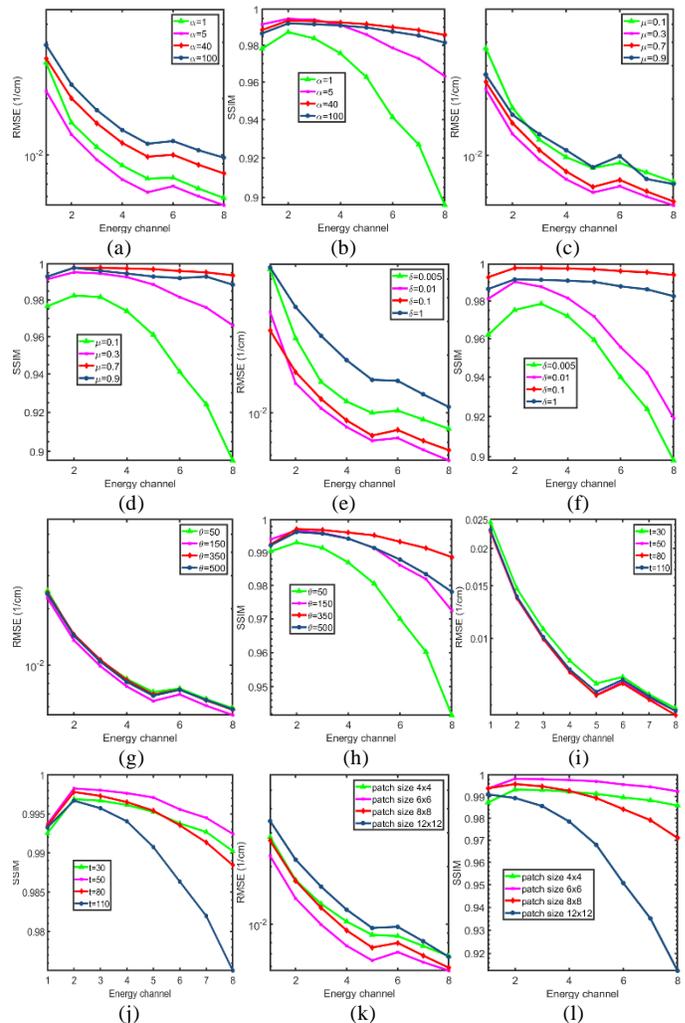

Figure 6. Parameters comparison in terms of RMSE and SSIM. (a) and (b), (c) and (d), (e) and (f), (g) and (h), (i) and (j), (k) and (l) are the RMSEs and SSIMs with different setting of the parameter $\alpha$, $\mu$, $\delta$, $\theta$, t and patch size, respectively.

### 4). Convergence and Computational Cost

There are two regularization terms in the KBR model, i.e., sparsity constraint of core tensor and low-rank property of tensor unfolding. In this study, the L$_0$-norm and nuclear norm are employed to respectively enhance the sparsity and low-rank properties. The optimization convergence is difficult to analyze. In addition, the L$_0$-norm minimization of core tensor coefficients is a nonconvex optimization problem, which also makes it more difficult to analyze the convergence. Alternatively, we only numerically investigate the convergence of the NLCTF method. Fig. 7 shows the averaged RMSE and PSNR values among all energy channels *vs.* iteration number. Since the projection datasets are corrupted by Poisson noise, the PSNR values of SART increase rapidly and then drop off slowly [7]. The RMSEs of all optimization methods are strictly decreasing with respect to iteration number and finally converge to a stable level. Particularly, the NLCTF can obtain a good solution with the smallest RMSE or a highest PSNR, followed by the SSCMF, $L_0$TDL, TV+LR and TV.

Regarding the computational cost, the NLCTF method is divided into two major procedures: data fidelity term update and regularization constraint. The backprojection reconstruction step is necessary for all the iteration algorithms and different regularization terms correspond to different computational costs. In this study, all the source code are programmed by Matlab (2017b) on a PC (8 CPUs @3.40GHz, 32.0GB RAM, Intel(R) HD Graphics 530). Table III summarizes the required time for one iteration for all algorithms.

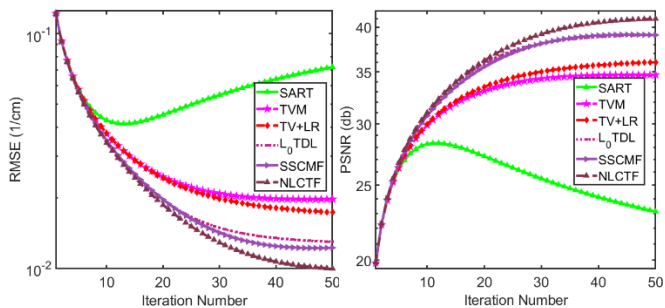

Figure 7. Convergence curves in terms of RMSEs and PSNRs.

Table III Computational costs of all reconstruction methods (unit: s).

| Methods | SART | TV | TV+LR | $L_0$TDL | SSCMF | NLCTF |
|---|---|---|---|---|---|---|
| Data term | 114.58 | 114.58 | 114.58 | 114.58 | 114.58 | 114.58 |
| Regularizer | 0 | 1.64 | 0.52 | 32.78 | 38.23 | 327.53 |

*3.B. Preclinical Mouse Study*

A mouse was scanned by a MARS micro spectral CT system including one micro x-ray source and one flat-panel PCD. The distances between the source to the PCD and object are 255 mm and 158 mm, respectively. The PCD horizontally includes 512 pixels and its length is 56.32 mm, resulting in an FOV with a diameter of 34.69 mm. Gold nanoparticles (GNP) are injected into the mouse as contrast agent. Because the PCD only contains two energy bins, multiple scans were performed to obtain 13 energy channels for 371 views with a cost of increased radiation dose. This can help to increase the stability of material decomposition and quality of decomposed basis material images. To compare all the reconstruction algorithms, the projections for the central slice are extracted and employed in this experiment. The size of each reconstructed channel image is 512×512.

Fig. 8 shows the reconstructed and gradient images of three representative energy channels (1st, 9th and 13th). The 1st row of images in Fig. 8 are reconstructed by the SART. Because they are corrupted by severe noises and lose most of details and fine structures, it is difficult to distinguish small bony details and soft tissues. The image quality of TV and TV+LR methods are improved (2nd and 3rd rows). From Fig. 8, we can see that the $L_0$TDL and SSCMF methods (4th and 5th rows) have significant advantages in recovering fine structures as well as preserving image edges than the TV and TV+LR results. Compared with the results of the proposed NLCTF, the capability of $L_0$TDL and SSCMF is weaker in preserving edges and recovering image features.

Here, ROIs indicated by "A" and "B" are extracted to demonstrate the aforementioned advantages of the NLCTF method. The magnified version of ROIs "A" and "B" are also given in Fig. 8. It can be seen that the bony profile in ROI "A" cannot be distinguished in the SART, TV and TV+LR results, especially in high energy bins (i.e., 9th and 13th). Compared with the TV and TV+LR methods, the $L_0$TDL can provide clearer bony profiles. However, the connected bone is broken in all channels in the $L_0$TDL results and high energy channels (such as 13) in the SSCMF results, which can be observed in the gradient images of ROI "A". From the reconstructed and gradient images of ROI "A", we can see that the proposed NLTDL can accurately recover the broken bony structure (e.g. the bone structure labelled by arrow "1"). For the ROI "B", the shape of bone is severely distorted in the images reconstructed by the SART and TV methods. The noise and blocky artifacts comprise the TV+LR final results. The shape of bony structures and image edges are also corrupted in the $L_0$TDL and SSCMF results, which can be seen from the location indicated by the arrow "2". Compared with all the competitors, the NLCTF can provide more accurate image edges and bony shapes.

A more complicated ROI "C" is extracted to further demonstrate the advantages of the NLCTF method. Fig. 9 shows the magnified ROI "C" of the $L_0$TDL, SSCMF and NLCTF methods. From Fig. 8, one can see that the SART, TV and TV+LR have a poor performance, and their results are omitted in Fig. 9 to save space. The image edge indicated by the arrow "3" is blurred, and it is hardly observed in all comparisons (i.e., $L_0$TDL and SSCMF methods). However, in the NLCTF results, one can easily see the sharp image structure edge and the image gap between two ribs. This is also verified by the gradient images. If we use the $L_0$TDL and SSCMF results in Fig 9, one may mistakenly conclude that the image structure "4" is broken and then lead to a wrong inference about the structure of the mouse. However, it is not difficult to infer that the image structure "4" is always continuous based on the NLCTF results. Again, the NLCTF may give more accurate structural information than other competitors. The image structure "5" reconstructed by the SSCMF is blurred and image edge quality is degraded, and the profile reconstructed by the $L_0$TDL method is slightly distorted in high energy bins. In contrast, the NLCTF method can avoid these drawbacks and enhance the ability of edge preservation. In term of image structure "6", the SSCMF and NLCTF methods can provide similar high quality images. However, the edge is blurred in the $L_0$TDL results, and this can be confirmed by the corresponding gradient images.

Better reconstructed image quality can benefits the basis material decomposition. Fig. 10 shows three decomposed basis materials. Regarding the decomposed bone results, ROIs indicated by "D" and "E" are extracted and magnified. From the magnified "D", a gap indicated by the arrow "7" can be easily observed in our proposed NLCTF result. However, it disappears in other competitors. The bony structure within the ROI "E" is always continuous for the SSCMF and NLCTF, while it is broken in the SART, TV, TV+LR and $L_0$TDL results. As for the decomposed soft tissue, the image structure, indicated by the arrow "8" in ROI "F", is well reconstructed by the NLCTF method, and image edges are clear than those obtained by other reconstruction methods. Besides, the gap between two bony profiles is much clearer than the SART, TV, TV+LR, $L_0$TDL and SSCMF methods. From the magnified ROI "G", it can be seen that the image structure reconstructed by the NLCTF technique provides sharper image edges. This point can be verified by the region around the arrow "9". For the iodine contrast agent decomposed results, the $L_0$TDL, SSCMF and NLCTF methods can obtain similar accuracy with clear image edges.

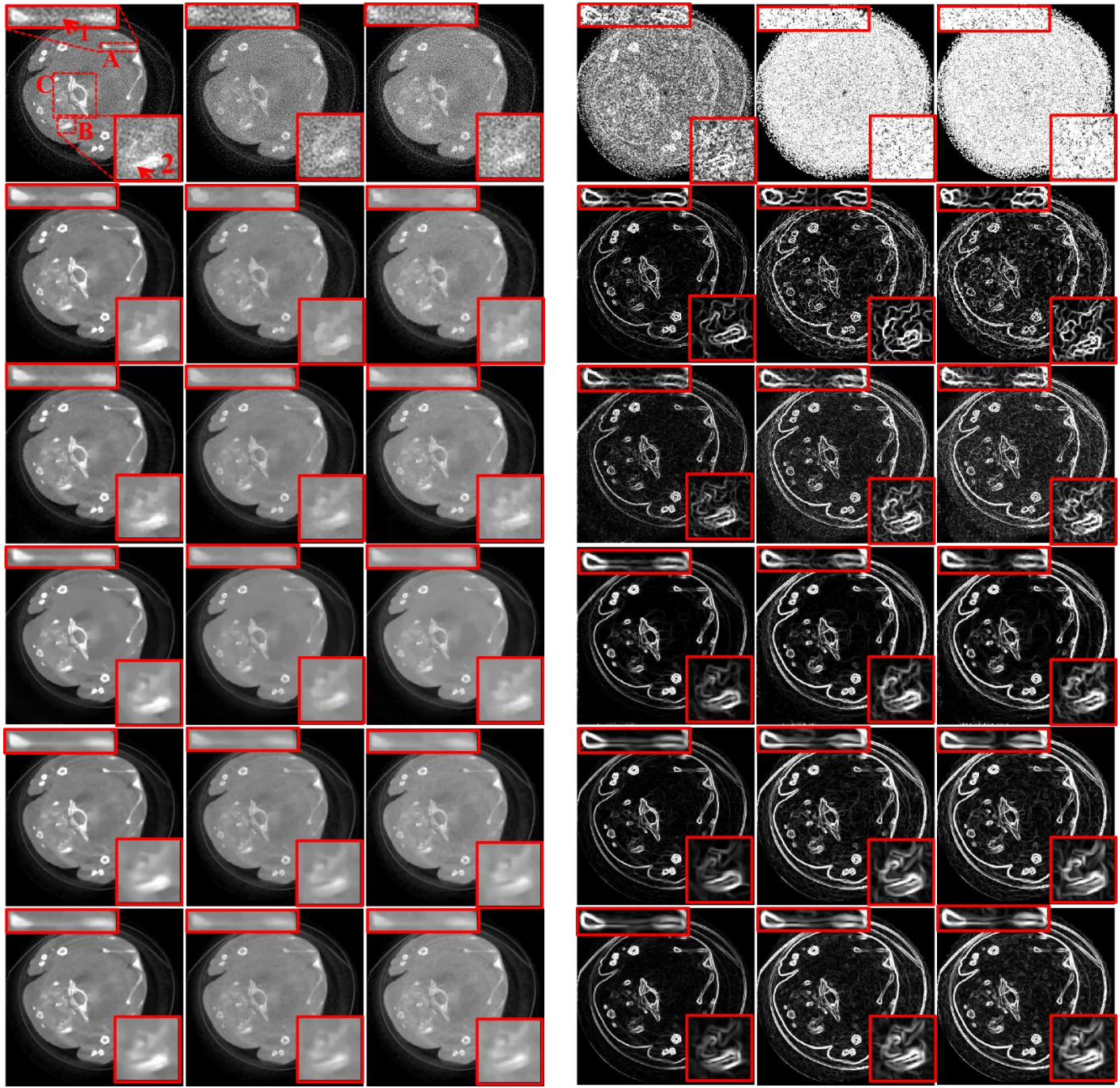

Figure 8. Preclinical mouse study results. The left panel shows the reconstructed original images from $1^{st}$, $9^{th}$ and $13^{th}$ (from left to right) energy bins and the right panel shows the corresponding gradient images. The $1^{st}$ to $6^{th}$ rows are reconstructed by using the SART, TV, TV+LR, $L_0$TDL, SSCMF and NLCTF methods, respectively. The display windows for $1^{st}$ to $6^{th}$ columns are [0, 0.8] cm$^{-1}$, [0, 0.8] cm$^{-1}$ [0, 0.8] cm$^{-1}$, [0, 0.08] cm$^{-1}$, [0, 0.04] cm$^{-1}$ and [0, 0.04] cm$^{-1}$, respectively.

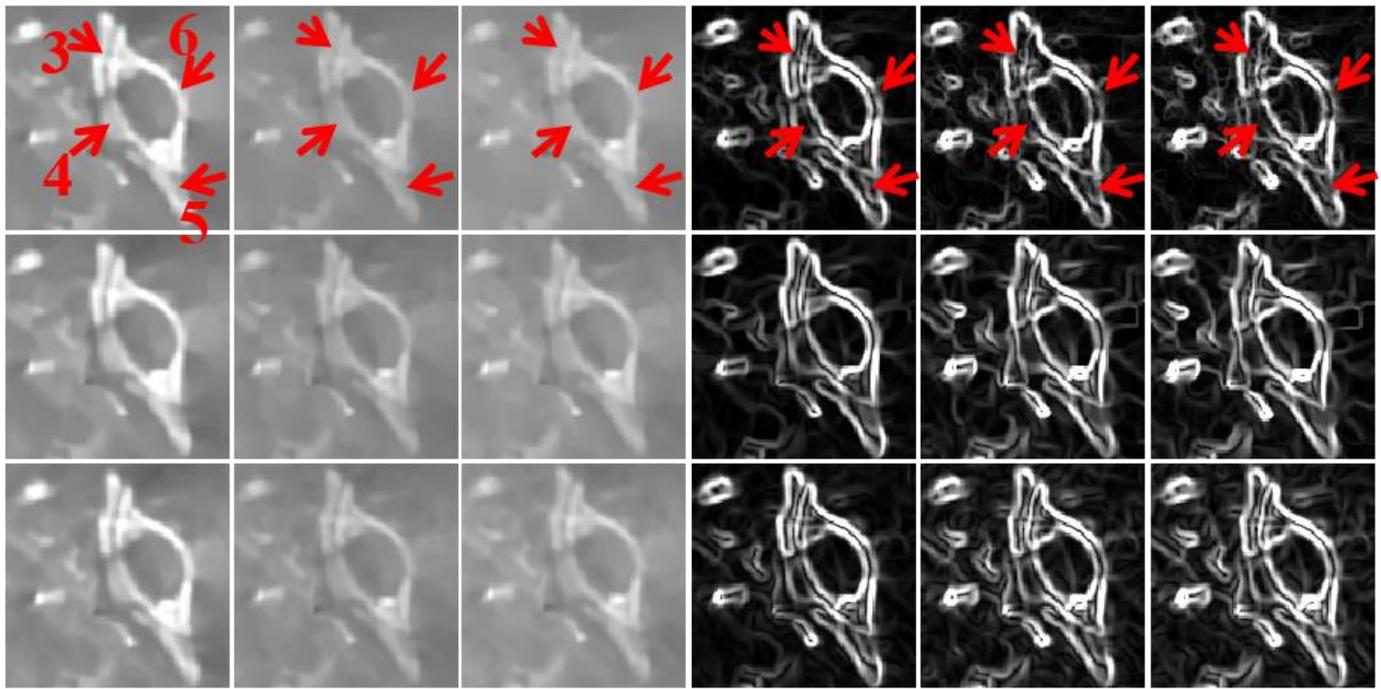

Figure 9. The magnified ROI "C" in Fig. 8 of the $L_0$TDL, SSCMF and NLCTF methods.

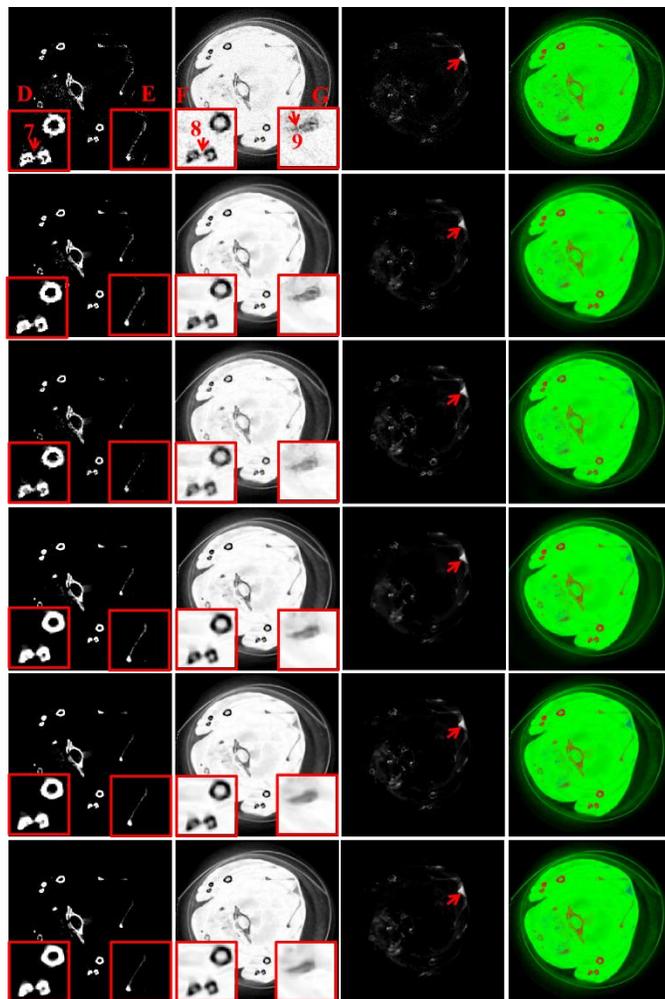

Figure 10. Decomposed basis materials from Fig. 8. The 1st to 3rd columns represent bone, soft tissue, and GNP. The 4th column images are the corresponding color rendering, where red, green and blue represent bone, soft tissue and GNP. The display windows for the 1st to 3rd columns are [0.1, 0.5], [0, 1] and [0, 0.5] cm$^{-1}$, respectively.

## IV. DISCUSSIONS AND CONCLUSIONS

To reconstruct high quality spectral CT images from noisy projections, the NLCTF method is proposed and developed. The NLCTF can sufficiently explore the low-rank property among the spatial-spectral space and image-self similarities by formulating small 3D cubes. Compared with the formulation of 4D group in the SSCMF, such 3D cubes can both fully encode the image spatial information and reduce memory load. Different energy bins correspond to different image contrast resolutions and noise levels. Specifically, higher energy channels have lower contrast resolutions so that the finer image details are difficult to distinguish. However, the noise levels of higher energy channels are lower than those obtained from lower energy channels. In contrast, the spectral images of lower energy channels have higher attenuation coefficients and contrast resolutions with larger noise levels than higher energy channels. The unfolding operations along every direction of the tensor $\mathcal{X}$ may be beneficial to improve the contrast resolutions in both higher and lower energy channels by reducing the noise. Thus, compared with the SSCMF method, NLCTF employs an advanced tensor factorization technique (KBR) to decompose the formulated 3D cubes rather than hard-thresholding and collaborative filtering operating on the 4D group in the SSCMF method. In this way, more image details and structures can be preserved in the final results.

To further validate the advantages of 3D cubes rather than 4D groups, Fig. 11(a) shows the nonlocal patch-based T-RPCA (NL-T-RPCA) results [36]. To the best of our knowledge, the cerebral perfusion CT mainly focuses on reconstructing both dynamic and static structures simultaneously. How to remove the motion artifacts with clear image structures is the biggest obstacle in practical application for cerebral perfusion CT. The success of NL-T-RPCA [36] is that it treats cerebral perfusion CT images as a low-rank component (static structures) and a sparsity component (dynamic structures). Then the KBR regularization and tensor-based total variation (TTV) regularization are employed to characterize the corresponding spatial–temporal correlations (low-rank) and spatial–temporal varying component (sparsity), respectively. This corresponds to the PRISM model reported by H. Gao et al. in 2011 [11]. Regarding the spectral CT, multi-energy projections are collected from the same object using different energy windows. Because the spectral CT images have different attenuation coefficients while sharing the same structures and details among different energy channels, it is more difficult to model the sparsity of the spectral-image. In addition, because there are only one or two materials within a small patch, it is appropriate to employ the KBR to realize small formulated 3D cubes. Here, the sparsity term in the NL-T-RPCA model is eliminated. It can be seen from Fig. 11(a) that the finer structures indicated by arrows are still smoothed and the image edges are hardly observed, which are similar to the SSCMF results.

Compared with the $L_0$TDL methods, the NLCTF can also obtain better results. To further demonstrate this point, the reconstructed mouse results using the TDL techniques are given in Fig. 11(b). Those results are the same as in [25]. From Fig. 11(b), it can be seen that the small structures indicated by arrows are broken, and the details cannot be observed. However, those structures can be clearly observed in our NLCTF results. This point has been mentioned in the above sub-section.

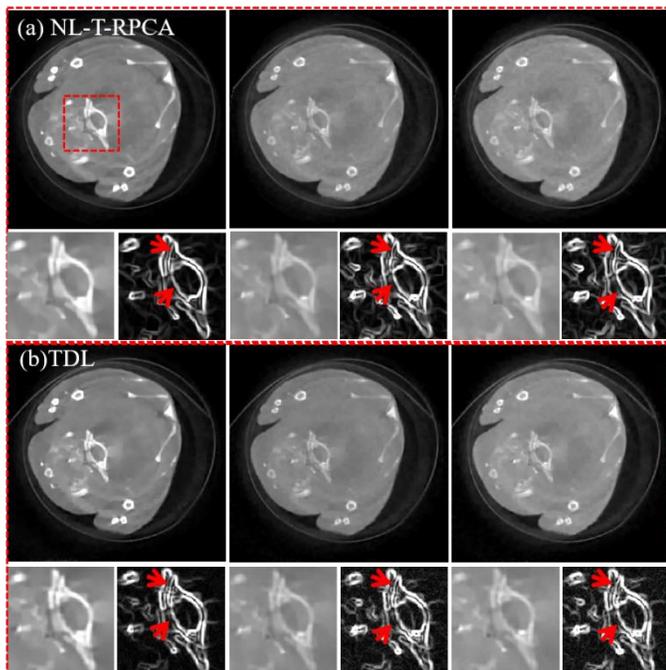

Figure 11. Reconstructed images of three representative energy channels ($1^{st}$, $9^{th}$ and $13^{th}$) using the NL-T-RPCA and TDL techniques. The $1^{st}$ and $3^{rd}$ rows shows the original images, and the $2^{nd}$ and $4^{th}$ rows show the original and gradient images of a magnified ROI.

While the NLCTF algorithm has an outstanding performance for spectral CT reconstruction, there are still some open problems in practical applications. First, there are several parameters in the NLCTF model which need to be selected for different objects, such as $\alpha$, $\delta$, $\theta$, $\mu$, *patch size*, *similar patch number, etc*. In this study, the parameters are carefully selected and optimized by comparing the values of different metrics. However, there may be no reference in practice. For that case, the final results are picked up based on our experiences, which may be inappropriate. Thus, it is important to develop a strategy for selecting good results with combining deeply theoretical analysis and extensive experiments in the future. Second, the NLCTF needs larger computational cost than the SSCMF, $L_0$TDL, TV+LR and TV. This can be speededup by the GPU techniques. Third, the proposed NLCTF currently focuses on fan-beam rather than cone-beam geometry. To generalize it to cone-beam geometry, a $4^{th}$-order low-rank tensor rather than the cube should be formulated.

In summary, the KBR measure is employed to decompose the non-local low-rank cube-based tensor to fully explore the similarities among spatial-spectral space, and the split-Bregman method is employed to solve the NLCTF model to obtain the optimized solution. Both simulation and preclinical experiments validate and demonstrate the outperformances of our proposed NLCTF reconstruction method.

**Acknowledgement**: The authors are grateful to Mr. Morteza Salehjahromi and Joshua Lojzim at UMass Lowell for their help on proofreading and language editing.